\documentclass[sigconf]{acmart}

\usepackage{amsmath,amsfonts}
\usepackage{algorithmic}
\usepackage{graphicx}
\usepackage{textcomp}
\usepackage{xcolor}
\definecolor{lightgray}{gray}{0.9} 
\usepackage{colortbl}

\usepackage{float}
\usepackage{multirow}
\usepackage{caption}

\usepackage{subcaption}

\usepackage{wrapfig}
\usepackage{enumitem}

\usepackage{amsthm}

\usepackage{etoolbox}

\newcommand{\N}{\texttt{GenHAR}}

\newcommand{\courier}{9,985}

\newcommand{\still}{1.84 billion}   
\newcommand{\walking}{215 million}  
\newcommand{\up}{54.8 million}  
\newcommand{\down}{39.8 million} 
\newcommand{\all}{2.15 billion} 

\copyrightyear{2026}
\acmYear{2026}
\setcopyright{cc}
\setcctype{by}
\acmConference[KDD '26]{Proceedings of the 32nd ACM SIGKDD Conference on Knowledge Discovery and Data Mining V.1}{August 09--13, 2026}{Jeju Island, Republic of Korea}
\acmBooktitle{Proceedings of the 32nd ACM SIGKDD Conference on Knowledge Discovery and Data Mining V.1 (KDD '26), August 09--13, 2026, Jeju Island, Republic of Korea}
\acmPrice{}
\acmDOI{10.1145/3770854.3783921}
\acmISBN{979-8-4007-2258-5/2026/08}

\begin{document}

\title{RealHAR: Real-time On-device Human Activity Recognition System by Modeling Signal Frequency}

\title{Real-time Frequency-Domain Transformer for On-Device Human Activity Sensing}

\title{Accelerating Attention-based DNN for Practical Real-time Human Activity Recognition}

\title{GenHAR: Generalizing Cross-domain Human Activity Recognition with Frequency Learning}
\title{GenHAR: Generalizing Cross-domain Human Activity Recognition with Efficient Frequency Attention}
\title{GenHAR: Generalizing Cross-domain Human Activity Recognition for Last-mile Delivery}



\author{Zhiqing Hong}
\affiliation{%
  \institution{HKUST (GZ)}
  \city{Guangzhou}
  \country{China}
}
\email{zhiqinghong@hkust-gz.edu.cn}

\author{Zelong Li}
\affiliation{%
  \institution{JD Logistics}
  \city{Beijing}
  \country{China}
}
\email{lizelong11@jd.com}

\author{Xiubin Fan}
\affiliation{%
  \institution{Beijing Normal–Hong Kong Baptist University}
  \city{Zhuhai}
  \country{China}
}
\email{fxiubin@gmail.com}

\author{Guang Yang}
\affiliation{%
  \institution{Rutgers University}
  \city{Piscataway}
  \country{USA}
}
\email{gy121@cs.rutgers.edu}

\author{Baoshen Guo}
\authornote{Baoshen Guo is the corresponding author.}
\affiliation{%
  \institution{SMART Center, MIT}
  \city{Singapore}
  \country{Singapore}
}
\email{baoshen@mit.edu}

\author{Haotian Wang}
\affiliation{%
  \institution{JD Logistics}
  \city{Beijing}
  \country{China}
}
\email{wanghaotian18@jd.com}

\author{Tian He}
\affiliation{%
  \institution{JD Logistics}
  \city{Beijing}
  \country{China}
  }
\email{tim.he@jd.com}

\author{Desheng Zhang}
\affiliation{%
  \institution{Rutgers University}
  \city{Piscataway}
  \country{USA}
  }
\email{desheng@cs.rutgers.edu}

\renewcommand{\shortauthors}{Zhiqing Hong et al.}

\begin{abstract}
Human Activity Recognition (HAR) has shown remarkable effectiveness in various applications, such as smart healthcare and intelligent manufacturing. 
However, a major challenge faced by HAR is the distribution shift across different sensor data domains, which often leads to decreased performance when deployed for real-world applications. 
To address this issue, this paper introduces \N, a novel framework designed to mitigate the domain gap by learning domain-invariant sensor representations.
\N\ aims to enhance the generalization capabilities of HAR on target domains purely with data from the source domain. 
The key novelty of \N\ lies in two aspects. 
Firstly, \N\ tokenizes sensor data and learns correlations among frequency sensor channel dimensions to improve the robustness of HAR models. 
Secondly, \N\ improves the efficiency via selective masking and an efficient attention mechanism. 
We conduct a systematic analysis of \N\ by comparing it with state-of-the-art HAR methods on real-world human activity datasets. 
Results show that \N\ outperforms state-of-the-art methods by 9.97\% in accuracy, and reduces Floating Point Operations by 6.4 times. 
Moreover, we deploy \N\ at a leading logistics company in 4 cities, and have detected \all\ real-time activities. 
We release our code at: \href{https://github.com/Sensor-Foundation-Model/GenHAR}{https://github.com/Sensor-Foundation-Model/GenHAR}.

\end{abstract}





\settopmatter{printacmref=true} 



\settopmatter{printfolios=false}
\maketitle

\section{Introduction}

\begin{figure}[t]
\centering
\includegraphics[width=3.3in]{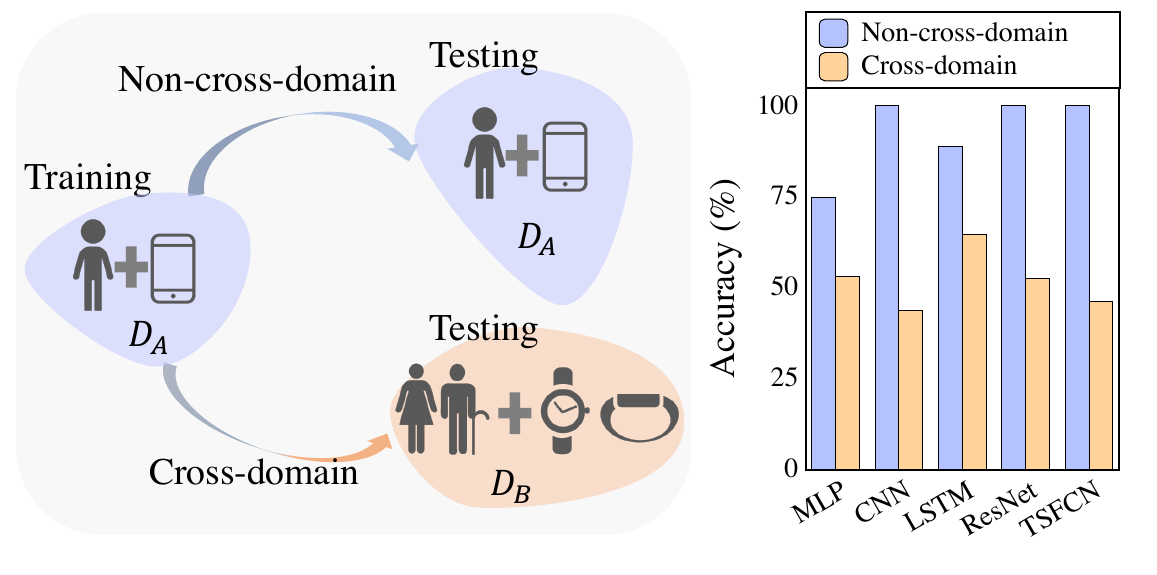}
\vspace{-6pt}
\caption{Illustration of performance degradation in on-device cross-domain HAR, where models are trained with source domain ($D_A$) and tested with the target domain ($D_B$).}
\label{fig:motivation}
\vspace{-6pt}
\end{figure}

In recent years, Human Activity Recognition (HAR) using Inertial Measuring Unit (IMU) has attracted increasing attention from the data mining community, given its widespread applications such as healthcare monitoring~\cite{ubicomp19_har_nursing} and manufacturing efficiency enhancement~\cite{ubicomp19_har_factory,ubicomp20_har_factory}.
However, one major challenge for HAR is the domain discrepancy between the training and testing data, 
This problem is also known as \textit{cross-domain HAR}~\cite{cross_dataset_ubicomp22_jindong,cross_domain_jindong}, leading to a marked decline in performance~\cite{cross_dataset_ubicomp22_jindong}.
For example, activity data gathered from one individual may significantly differ from another's due to differences in sensor positioning, and individual behavioral patterns, etc. 
Fig.~\ref{fig:motivation} presents two typical settings of the HAR problem.
In the non-cross-domain setting, both the training and testing data originate from the same domain ($D_A$). 
Conversely, in the \textit{cross-domain} setting, the training and testing data come from different domains ($D_A$ and $D_B$, respectively). 
We observe that the performance of popular HAR models declines significantly (shown in Fig.~\ref{fig:motivation}) in the cross-domain setting, demonstrating the practical challenge of cross-domain HAR. 
Therefore, we aim to generalize HAR across domains for real-world applications, especially for last-mile delivery.

Existing work generally addresses cross-domain HAR by designing \textit{domain adaptation} methods, which enhance the generalization capabilities of models trained on source domain data by utilizing data from the target domain~\cite{cross_user_aaai23,ubicomp22_har_cross_domain}.
A recent advancement involves using unlabeled data from target domains~\cite{unihar}.
However, obtaining data from the target domain is not always feasible, and the process of collecting such data along with its associated labels can be both time-consuming and costly. 
As a result, some studies have focused on leveraging only source domain data to improve model generalization capabilities, i.e., \textit{domain generalization}~\cite{wang2022generalizing}. 
There are different methods for domain generalization such as data augmentation~\cite{DG_data_augmentation}, knowledge distillation~\cite{DG_knowledge_distill,DG_knowledge_distill_jindong}, and self-supervised learning~\cite{kim2022broad,kim2021selfreg}.
Nevertheless, most of these methods were designed for other modalities like images~\cite{kim2021selfreg}, without fully exploiting the characteristics of IMU sensor data. 
Only a few studies have focused on improving the generalization capability of HAR totally with data from only source domains~\cite{ddlearn_kdd23,cross_dataset_ubicomp22_jindong}.
Due to the complexity and dynamics of IMU data, most studies have not achieved optimal performance when multiple distribution shifts exist simultaneously~\cite {cross_dataset_ubicomp22_jindong}. 
Moreover, the increasing parameter size of HAR models~\cite{survey_har_yunhao} brings challenges to edge devices with low computational resource budgets.

We argue that IMU data is different from general time series like finance~\cite{finance_prediction}, weather~\cite{weather_predict_science}, and COVID~\cite{covid_predict}, in terms of the data generation process, multi-sensor and multi-channel correlation, and sensor property heterogeneity. 
Accurate cross-domain HAR mainly experiences the following challenges: 

\begin{itemize}[leftmargin=*]
    \item \textit{(i) Uncertainty of distribution shift.} The distribution between the source and target domains varies significantly. 
    Different habits generate IMU signals with diverse patterns. 
    A variety of noises, including sensor noise, human movement noise, etc., impact the readings from IMU sensors. Moreover, multiple discrepancies could simultaneously exist between the source domain and the target domain, such as user groups and device placements~\cite{crosshar}. 
    \item \textit{(ii) Unavailability of target domain data.} In the real-world scenario, it is common that the data from target domains are unavailable during the model training phase. The model can only be trained with source domain data and is supposed to generalize well to unseen target domains. 
    \item \textit{(iii) Real-time requirement of mobile devices.} In practice, HAR models are supposed to run on mobile and edge devices with limited computation costs. Especially when the goal is real-time and long-term HAR on low-resource devices. Therefore, reducing computation costs while maintaining high recognition accuracy is crucial. 
\end{itemize}

To address these challenges, we design \N, a novel cross-domain HAR framework. 
(i) To address the first and the second challenges, we transform the IMU signals into the frequency space, rather than in the widely studied time space~\cite{survey_har_yunhao,imubert}, and we design a novel framework for learning only with data from the source domain. 
Firstly, by converting the temporal time series into the frequency space, we obtain a compact representation of the original multi-channel signal. 
Then, by focusing on the frequency domain, \N\ can learn patterns that are less sensitive to the domain shift.
Furthermore, unlike existing studies, we argue that the correlation between different sensor channels is critical for distinguishing different categories, as well as easy to be impacted by domain shift. 
Therefore, \N\ incorporates a novel sensor-wise self-attention module to capture the inherent correlations among sensor channels.
(ii) To address the third challenge, we consider efficiency as a constraint and reduce the computation significantly through two design techniques, i.e., calculate self-attention along the sensor channels rather than along the original sequence, and design a selective masking strategy to reduce the calculation in self-attention. 
In summary, this paper makes the following contributions:

\begin{itemize}[leftmargin=*]
    \item To our knowledge, this is the first work for cross-domain HAR by learning sensor attention in the signal frequency space. Compared to most existing studies, our work sheds light on cross-domain HAR without any training data from the target domains. 
    \item Technically, we design \N, a novel frequency-space Transformer architecture, learning generalizable IMU representation by capturing sensor correlations. Our model explores the possibility of learning generalizable representation fully from the frequency space. 
    \item Experimentally, we evaluate \N\ on 4 real-world datasets. Results show that \N\ outperforms the best baseline models by \textbf{9.97\%} in average accuracy. Moreover, \N\ is \textbf{43.2 times smaller} in self-attention computation, and \textbf{6.4 times smaller} in terms of overall Floating Point Operations (FLOPs) compared to the best baseline model. 
    \item Moreover, we deploy \N\ at a leading logistics company on \courier\ delivery couriers in 4 cities for delivery behavior understanding and map data mining. \N\ has detected \textbf{2.15 billion} activities in a month, including \still\ still, \walking\ walking, \up\ upstairs, and \down\ downstairs. 
\end{itemize}

\section{Background and Problem Formulation}
\label{sec:data-and-methoddology}

\subsection{HAR in Last-mile Delivery}
In last-mile delivery, couriers deliver E-commerce packages to customers across diverse urban environments. Accurately recognizing couriers' activities could support applications like workload-aware task assignment to improve welfare. 
HAR for couriers is a kind of cross-domain HAR, given the distribution shift across domains. 

\noindent
\textbf{(i) Distribution shift of IMU data}.
A "domain" in this scenario can be formally defined as \textit{a specific environment or context within which the sensor data is collected}. 
This includes the combination of factors such as specific sensor characteristics, sensor placement, device orientations, data collection protocol, and the individual characteristics of the user group from which the data is sourced. 
Given the sensitivity of IMU sensors, all these factors impact the sensor readings, which can mislead the HAR model and fail to capture domain-invariant activity-related features, finally decreasing the model performance.

\noindent
\textbf{(ii) Cross-domain HAR}.
According to the availability of data from target domains during the training phase, two approaches can address 
cross-domain HAR. 
\textit{Setting a: Domain adaptation}: training HAR models with data from both source and target domains. 
\textit{Setting b: Domain generalization}: training HAR models with data only from the source domain. 
We focus on \textit{Setting b} since it is more practical in real-world last-mile delivery scenarios.

\begin{figure}
\centering
\includegraphics[width=3.3in]{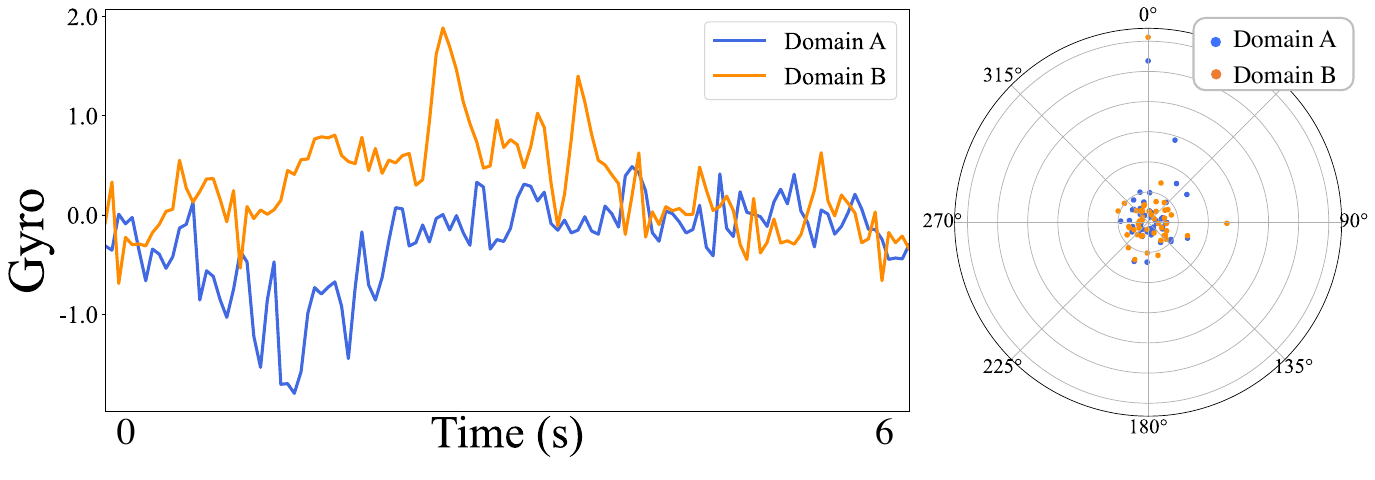}
\vspace{-6pt}
\caption{Frequency space has a smaller distribution shift than in time-space. \underline{Left}: sensor readings from one channel of the gyroscope sensor in the time-space. \underline{Right}: corresponding polar coordinates of Fourier features from Domain A and B.} 
\label{fig:motivation_time_fre}
\vspace{-6pt}
\end{figure}

\subsection{Motivations of Frequency Learning}

While existing methods can partially address the distribution shift problem, most of them focus on the time-space of IMU data~\cite{cross_dataset_ubicomp22_jindong,unihar,wang2022generalizing}. 
The fundamental reason that causes the distribution shift is that time features are easily impacted by random factors such as noises or subtle motion preferences. 
Therefore, two samples belonging to the same category might have very different time-space representations~\cite{imubert}. 
However, frequency features are more robust to time shifts and have the potential to provide domain-invariant features.
For example, as shown in Fig.~\ref{fig:motivation_time_fre}, both accelerometer and gyroscope, two components of IMU data, have great distribution shift across two persons in UCI~\cite{uci} and HHAR~\cite{hhar} datasets. 
However, in the frequency space, the Fourier features are most domain invariant, i.e., frequency features from source and target domains have similar distribution compared to time features. 
Existing studies have demonstrated that the divergence between source domains and target domains is the key that impacts the domain adaptation performance~\cite{ben2010theory,he2023domain}. 
Such closer distribution in frequency space can bring better performance on target domains~\cite{he2023domain}. 
Recently, in the computer vision community, studies have also shown the benefit to domain generalization brought by properly modeling frequency-domain features~\cite{fre_domain_generalization_cvpr,frequency_dg_explain_cvpr20}.
Therefore, learning in frequency space could potentially improve the domain generalization of HAR.

\subsection{Problem Definition}
\label{sec:problem_definition}

Let $\mathbf{X} = [X_1,X_2,...,X_M]$ $\in$ $\mathcal{X}$ $\subset$ $\mathbb{R}^{M \times T}$ represents an IMU sensor data with $M$ channels and $T$ time steps.
$X_m$ $\in \mathbb{R}^{T}$ is the $m$-th channel IMU data, and $y$ $\in$ $\mathcal{Y}$ $\subset$ $\mathbb{R}^{K}$ is the corresponding category for $\mathbf{X}$, $K$ is the total number of potential categories. 
Given $D_{source}$ $\in \mathbb{R}^{n_s \times M \times T}$ containing $n_s$ IMU samples, 
we aim to learn a model $\mathcal{M}(.,\Phi)$: $\mathcal{X}$ $\rightarrow$ $\mathcal{Y}$, which can generalize and perform well to unseen target dataset $D_{target}$ $\in \mathbb{R}^{n_t \times M \times T}$ with $n_t$ samples, which is not available in the training process. 
Note that $\mathbf{X}$ can be a multi-sensor and multi-channel time series, i.e., $M = S \times C$, where $S$ is the number of sensors of IMU and $C$ is the number of channels for each sensor, each sensor has the same number of channels.

\section{\N\ System Design}
\label{sec:framework}

\subsection{System Overview}

\begin{figure}
\centering
\includegraphics[width=3.3in]{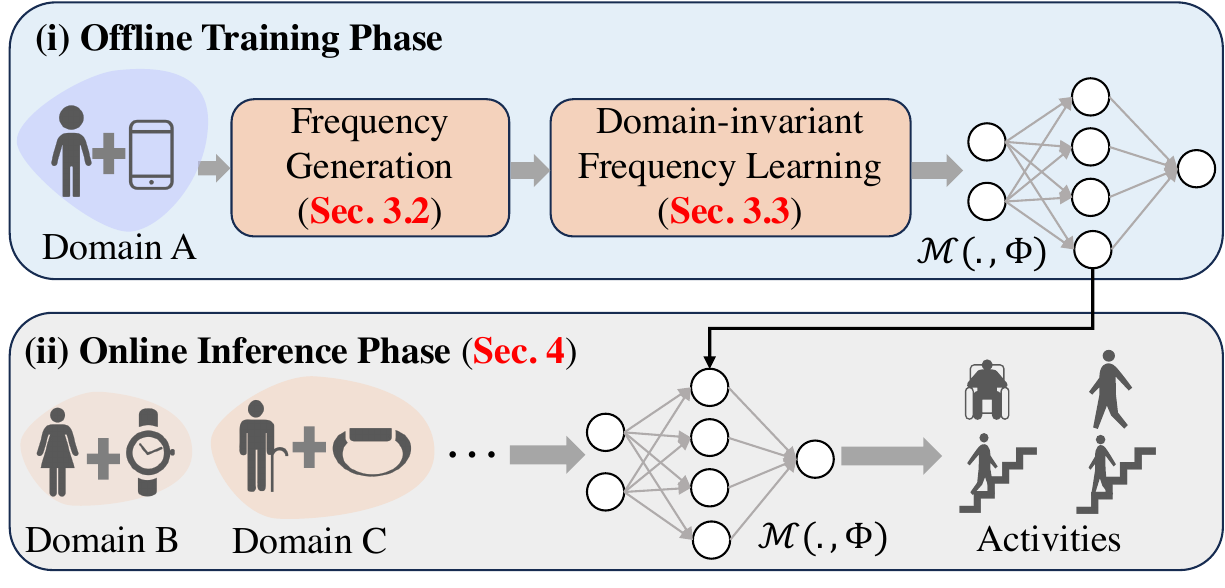}
\vspace{-6pt}
\caption{Overall framework of \N. \N\ consists of an (i) Offline Training Phase and an (ii) Online Inference Phase. 
}
\label{fig:overall_framework}
\vspace{-6pt}
\end{figure}

As shown in Fig.~\ref{fig:overall_framework}, 
\N\ consists of the following two phases.

\noindent
\textbf{(i) Offline Training Phase.}
In the Training Phase, only data from the source domain is utilized to train a HAR model $\mathcal{M}(.,\Phi)$. The Training Phase consists of two key components: 
\textit{(1) Frequency Generation} (Sec.~\ref{sec:frequency_generation}). First, we transform the raw IMU data from the time into the frequency space. As a result, we enable our model to extract more robust IMU representations to generalize to unseen target domains. 
\textit{(2) Domain-invariant Frequency Learning} (Sec.~\ref{sec:representation_learning}). Then, we design a frequency transformer model to learn transferable IMU representations. A channel-wise self-attention module is designed to capture the correlation among different sensor channels. 

\noindent
\textbf{(ii) Online Inference Phase.}
After $\mathcal{M}(.,\Phi)$ is generated, 
in the Online Inference Phase, data from target domains are utilized to evaluate the cross-domain performance of $\mathcal{M}(.,\Phi)$. Then, $\mathcal{M}(.,\Phi)$ is also evaluated by deploying it on mobile devices. 
The details are in Sec.~\ref{sec:evaluation} and Sec.~\ref{sec:deployment}.

\subsection{Frequency Generation}
\label{sec:frequency_generation}
In this subsection, we introduce the details of frequency generation and the motivation for generating amplitude rather than phase.

\subsubsection{Preliminary of FFT}
Discrete Fourier Transform (DFT) is a fundamental tool to analyze the frequencies of various signals and has wide applications in sequence data processing, such as signal processing, audio analysis, and many areas of engineering.
Fast Fourier Transform (FFT)~\cite{fft} is an efficient algorithm to compute DFT to obtain the frequency information.

\subsubsection{Frequency extraction of IMU data}

Given a multivariate IMU data $\mathbf{X} = \{X_1, X_2, \ldots, X_M\}$ $\in$ $\mathbb{R}^{M \times T}$ with $M$ channels and $T$ time steps, where $X_m(t)$ represents the signal in the $m$-th channel at time \( t \). 
For each channel $m$, FFT is applied to obtain the frequency components:

\begin{equation}\small
    \hat{X}_m(k) = \mathcal{F}\{X_m(t)\} = \sum_{t=0}^{T-1} X_m(t) \cdot e^{-\frac{2\pi j}{N} k t}, \quad k = 0, 1, \ldots, T-1
\end{equation}
where $\mathcal{F}[\cdot]$ denotes the FFT. $\hat{X}_m(k)$ is the complex number representing the amplitude and phase of the frequency component at frequency index $k$ for the $m$-th channel, $j$ is the imaginary unit, and $T$ is the number of time steps in the IMU signal.

After applying the FFT to all channels of $\mathbf{X}$, we obtain a complex-valued IMU signal $\hat{\mathbf{X}}$ $\in$ $\mathbb{C}^{M \times T}$. 
Each element in $\hat{\mathbf{X}}$ can be represented as a combination of a real part $a_{m,k}$ and an imaginary part $b_{m,k}$ as: 
\begin{equation}
\label{eqn:comlex_num}
    \hat{X}_m(k) = a_{m,k} + ib_{m,k}
\end{equation}
where $\hat{X}_m(k)$ is a complex number in the frequency domain. \( a_{m,t} \) is the real part and \( b_{m,t} \) is the imaginary part. \( i \) is the imaginary unit  and satisfies \( i^2 = -1 \).

Note that, since $\mathbf{X}$ is a real-valued IMU data, the result of the FFT $\hat{\mathbf{X}}$ is symmetric, with the second half of the spectrum being the complex conjugate of the first half. 
Due to this symmetry, we cut off the second half of $\hat{\mathbf{X}}$ to reduce computation complexity and improve the model efficiency. 
Moreover, the first element of the $\hat{\mathbf{X}}$, often referred to as the "DC component," represents the average or mean value of the entire signal. 
It is normally not considered a meaningful frequency component and thus can be removed~\cite{xu2023fits}. 
We get $\tilde{\mathbf{X}}$ $\in$ $\mathbb{C}^{M \times \lfloor \frac{T}{2} \rfloor}$ as follows:

\begin{equation}
    \tilde{\mathbf{X}} = \hat{\mathbf{X}}[:,1:\tilde{T}], \quad \tilde{T} = \lfloor \frac{T}{2} \rfloor  
\end{equation}
where $\lfloor \frac{T}{2} \rfloor$ represents rounding down $\frac{T}{2}$, if \( L \) is odd, it would be \( \hat{X}_{m,1} \) to \( \hat{X}_{m,\frac{L-1}{2}} \).
Therefore, for each channel $m$, we get $\tilde{\mathbf{X}}_{m} = \left[ \hat{X}_{m,1}, \hat{X}_{m,2}, \ldots, \hat{X}_{m,\lfloor \frac{T}{2} \rfloor} \right]$.


\begin{figure}
\centering
\includegraphics[width=3.3in]{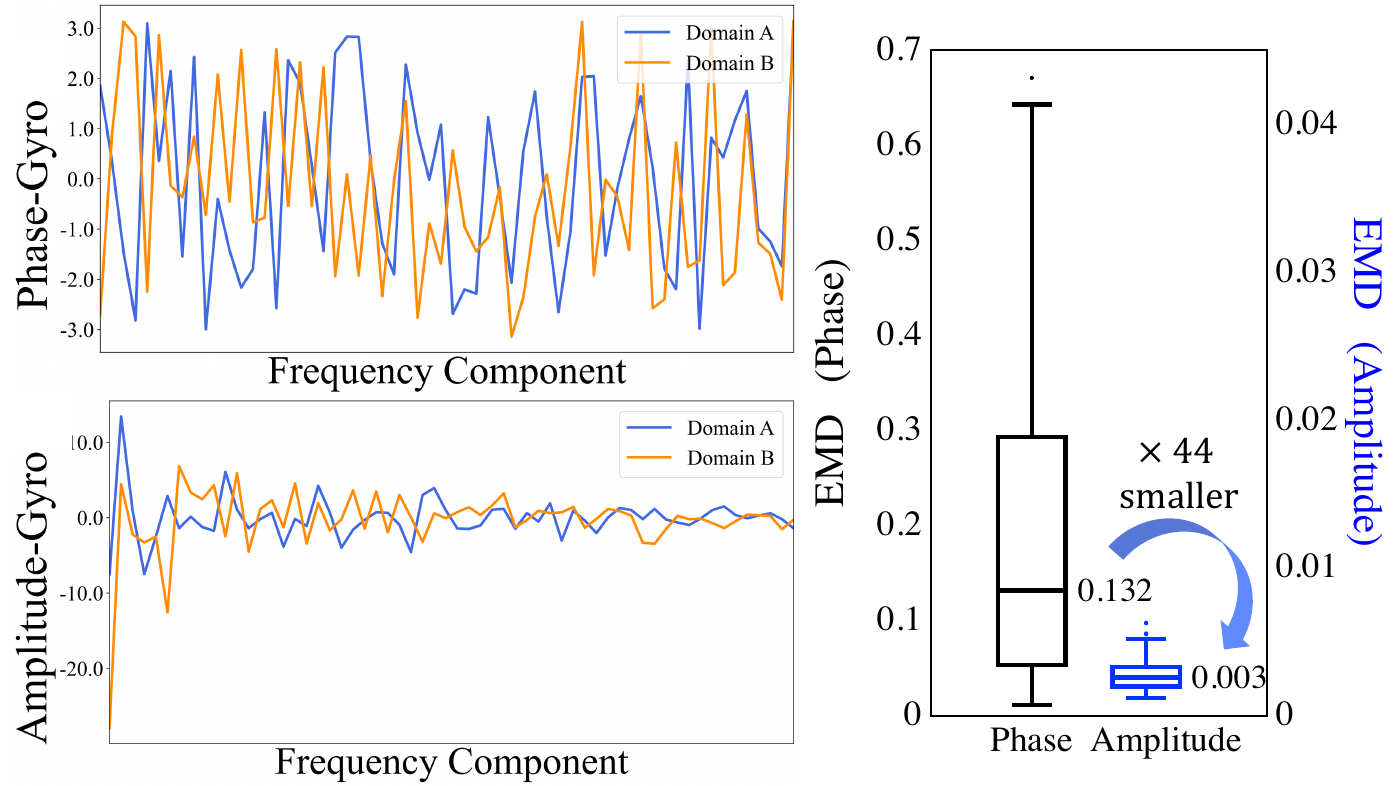}
\caption{Amplitude features are more domain-invariant than phase features. \underline{Left}: Phase features and corresponding amplitude features; \underline{Right}: Earth Mover's Distance between Domain A and Domain B of phase and amplitude.}
\label{fig:motivation_phase_amp}
\end{figure}

\subsubsection{Amplitude generation}
Unlike most existing studies utilizing the complex value or amplitude and phase information~\cite{xu2023fits,frequency_mlp_nips23,finding_order_nips23}, we propose to focus on amplitude in the frequency space for better generalization capability due to the following reasons. 
Firstly, amplitude possesses relative stability across various activity collection scenarios. 
This stability makes it easier to generate domain-invariant IMU representations. 
On the other hand, phase information is often sensitive to noise and device orientations.
As shown in Fig.~\ref{fig:motivation_phase_amp}, the phase features from different domains deviate from each other significantly. At the same time, the amplitude features remain relatively stable and consistent. 
Quantitatively, we calculate the earth mover distance of data from two domains, which is widely applied to measure the similarity between two distributions~\cite{emd_vldb}. 
A higher EMD value means a greater difference in distribution. 
For example, for the first channel of the accelerometer sensor, the EMD is 0.015 in the time-space, the EMD is 0.033 in the phase space, and the EMD is 0.003 in the amplitude space, respectively. 
When transformed into the amplitude space, the distribution gap between source and target domains decreases significantly. 
Therefore, we utilize pure amplitude information for cross-domain HAR.

In the complex domain as shown in Eqn.~\ref{eqn:comlex_num}, we can derive the amplitude for each channel $m$ at timestamp $t$ as: 

\begin{equation}
    |\hat{X}_{m,t}| = \sqrt{a_{m,t}^2 + b_{m,t}^2}
\end{equation}

Therefore, for $\tilde{\mathbf{X}}$ $\in$ $\mathbb{C}^{M \times \tilde{T}}$, we derive the IMU data $\mathbf{X}'$ $\in$ $\mathbb{R}^{M \times \tilde{T}}$ in amplitude space. Each sample is now represented by corresponding amplitude features with length $\tilde{T}$.

\subsection{Domain-invariant Frequency Learning}
\label{sec:representation_learning}
\begin{figure}[t]
\centering
\includegraphics[width=3.0in]{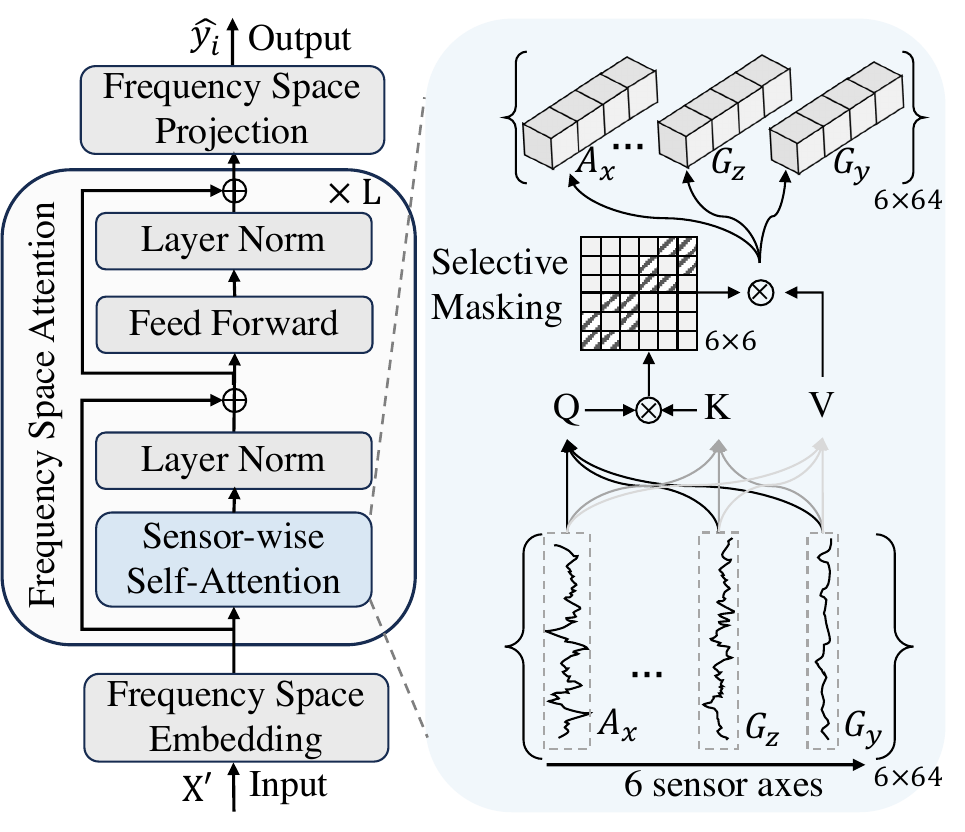}
\vspace{-6pt}
\caption{Details of domain-invariant frequency learning.}
\label{fig:har_model}
\vspace{-13pt}
\end{figure}

Once we have transformed time to the frequency space and obtained the amplitude representation, the next step is to learn the frequency representation for IMU data. 
Our domain-invariant frequency learning is shown in Fig.~\ref{fig:har_model}, 
consisting of three main modules, i.e., a Frequency Space Embedding, a Frequency Space Attention, and a Frequency Space Projection module.

\subsubsection{Frequency space embedding}

Given amplitude sequence $\mathbf{X}'$ $\in \mathbb{R}^{M \times \tilde{T}}$, we first obtain the amplitude embedding. 
\begin{equation}
\label{eqn:embedding}
    \mathbf{E} = \mathtt{AmpEmbedding}(\mathbf{X}')
\end{equation}
where $\mathtt{AmpEmbedding}$ is implemented by a trainable linear projection $\mathbf{W}$ $\in \mathbb{R}^{\tilde{T} \times d_{model}}$, following previous practices~\cite{Yuqietal-2023-PatchTST,imubert}. 
For each channel $m$ in $\mathbf{X}'$, we apply an embedding layer to represent each channel and finally get
$\mathbf{E}$ $\in \mathbb{R}^{M \times d_{model}}$, where $d_{model}$ is the hidden dimension.
$\mathtt{AmpEmbedding}$ learns the correlation among different frequency components in each channel. 

\subsubsection{Frequency space attention}
Based on the amplitude embedding, we design our representation learning component based on self-attention, which has been proven good at processing sequence data~\cite{wu2021autoformer,weather_predict_science,liu2021gated}. 
In the original Transformer architecture~\cite{transformer}, self-attention allows each position in the input channel to attend to all positions in the same channel to compute a channel representation. 
In our setting, each sensor captures a specific physical property, and each channel in one sensor provides information from a different angle. 
The accelerometer measures the linear acceleration, and the gyroscope measures the angular velocity~\cite{imu_princeple}. 
Therefore, we argue that the correlation among different channels is more important for distinguishing different categories. 
Moreover, in the cross-domain setting, such correlation information is more domain-invariant compared to the information in each sensor channel. 
Following this intuition, we design a novel sensor-wise attention method in the frequency space for cross-domain HAR.

\begin{figure}[t]
\centering
\includegraphics[width=3.0in]{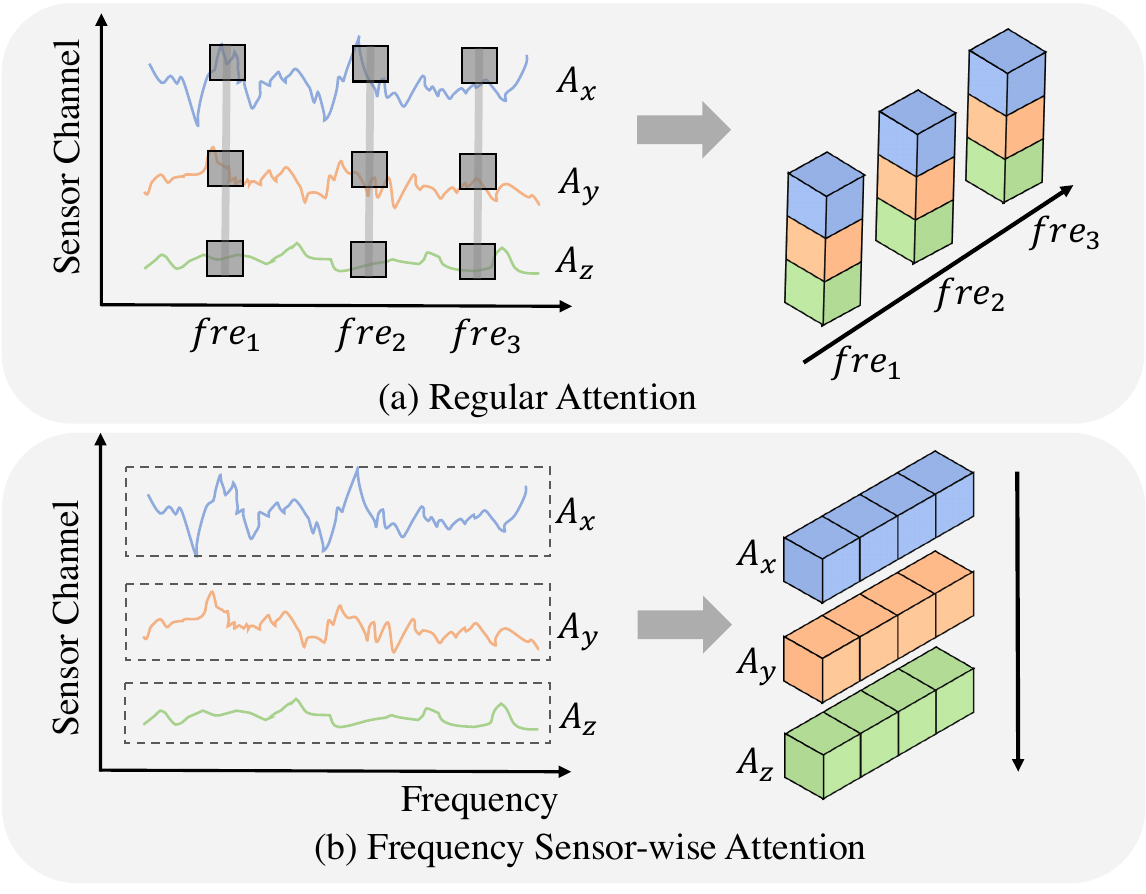}
\vspace{-6pt}
\caption{Sensor-wise self-attention improves efficiency.}
\label{fig:channel_attention}
\vspace{-6pt}
\end{figure}

\vspace{3pt}
\noindent
\textbf{Sensor-wise Self-attention.}
For each sensor channel, we take the whole sequence of amplitude components as one token, a basic granularity for conducting self-attention, similar to a \textit{word} in natural language processing~\cite{bert}.  
Then, we design self-attention along the sensor channel dimension, calculating attention across different IMU channels at the same position, allowing each channel to attend to all channels. 
We illustrate the details of sensor-wise attention in Fig.~\ref{fig:channel_attention}(b). Compared to normal attention in the frequency space as in Fig.~\ref{fig:channel_attention}(a), our sensor-wise attention calculates attention across all channels, while normal frequency attention calculates attention among different frequency components for each channel.


Based on frequency amplitude embedding $\mathbf{E}$ in Eqn.~\ref{eqn:embedding}, we derive three matrices $Q \in \mathbb{R}^{M \times d_k}$, $K \in \mathbb{R}^{M \times d_k}$, and $V \in \mathbb{R}^{M \times d_v}$ via linear transformation. 
$Q$, $K$, and $V$ are query, key, and value matrices. $M$ is the number of channels and $d_k$, $d_v$ are the dimensions of the key and value vectors. We define the self-attention as:
\begin{equation}\small
     \mathbf{H} = \texttt{SensorAttention}(Q, K, V) = \texttt{Softmax}\left(\frac{QK^T}{\sqrt{d_k}}\right)V
\end{equation}
$\mathtt{Softmax}$ is applied row-wise, i.e., along each sensor channel.  ($\sqrt{d_k}$) which is used for scaling.
$\texttt{Softmax}\left(\frac{QK^T}{\sqrt{d_k}}\right)$ is the correlation among all channels, i.e., attention map matrix, representing the sensor-wise correlation in the IMU signal. 
$\mathbf{H} \in \mathbb{R}^{M \times d_v}$ represents the matrix of attention-modified channel embeddings.
Note that even though a few studies focus on self-attention along different dimensions~\cite{liu2021gated,liu2023itransformer}, none of them has explored the correlation among sensor channels in the frequency space and the corresponding impact on cross-domain HAR.





\vspace{3pt}
\noindent
\textbf{Selective Masking.} 
Our sensor-wise self-attention captures the correlation among different sensor channels. 
For activity recognition, the correlations between different sensor channels can have very different importance. 
Therefore, sensor-wise attention may not effectively represent these variances in importance. 
Moreover, we can potentially reduce the computation by filtering attention calculations with low importance. 
Consequently, we further design a new selective masking mechanism for learning sensor correlations.

\begin{figure}[t]
\centering
\includegraphics[width=3.1in]{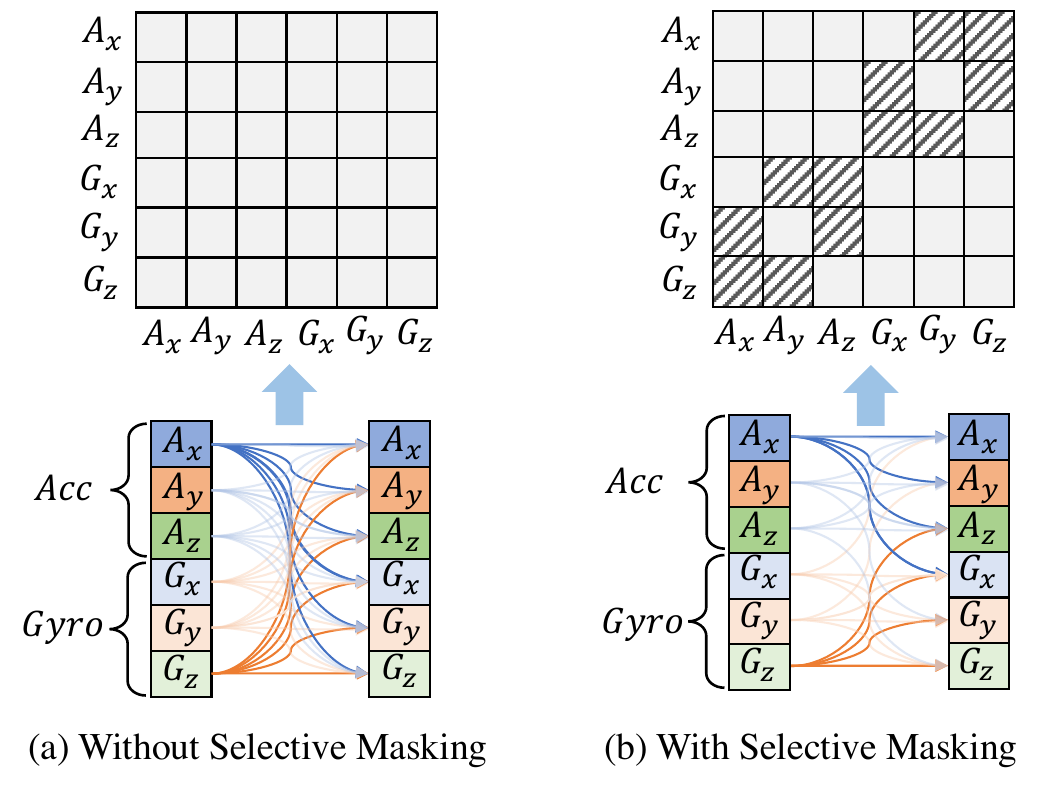}
\vspace{-6pt}
\caption{Self-attention without selective masking (a) and self-attention with selective masking (b).}
\label{fig:selected_attention}
\vspace{-6pt}
\end{figure}

This idea is introduced in Fig.~\ref{fig:selected_attention}. 
Specifically, we mask the correlation among different sensors at different channels. 
This is because the $Acc$ measures linear acceleration, while the gyroscope measures angular velocity, and their measurements in different directions do not have a direct physical connection~\cite{imu_princeple}.
For example, a change in $A_x$ typically does not directly affect the angular velocity in the y or z directions of the $Gyro$.
It can be noted that with selective masking, the computation is more efficient, reducing 33.3\% matrix multiplication in attention map calculation. 
Moreover, our selective masking mechanism can guide our model to concentrate only on intra-sensor correlations (i.e., $Acc$ and $Gyro$) and inter-sensor correlations (e.g., $A_x$ and $G_x$). 





\vspace{3pt}
\noindent
\textbf{Layer Normalization.} 
In our setting, we calculate self-attention along the channel dimension. 
Thus, we adapt layer normalization~\cite{layernorm} to normalize IMU data across positions for each channel, rather than across the feature dimension at each position separately~\cite{transformer}. 
Given $\mathbf{H}$ from the $\mathtt{ChannelAttention}$ module, we have: 
\begin{equation}\small
    \mathbf{A} = \texttt{LayerNorm}_a(\mathbf{E} + \mathbf{H})
\end{equation}
where $\mathbf{E}$ is the input to the self-attention.
The residual connection helps with the gradient flow during backpropagation.

\vspace{3pt}
\noindent
\textbf{FeedForward.}
The representation after $\texttt{LayerNorm}_a$ is further processed by a $\mathtt{FeedForward}$ module. 


\begin{equation}
    \mathbf{S} = \texttt{LayerNorm}_f(\mathtt{MLP}(\mathbf{A}) + \mathbf{A})
\end{equation}
where $\mathbf{S}$ $\in \mathbb{R}^{M \times d_{model}}$ is the output of the $\mathtt{FeedForward}$, and $d_{model}$ is the output dimension of the $\mathtt{MLP}$. 

\subsubsection{Frequency space projection}
Based on the representation derived from previous components, we further design a $\mathtt{SensorMixing}$ Layer to fuse the knowledge from different channel representations. 
Then, a linear layer with an activation function projects the representation into probabilities of different activity categories. 
Given $\mathbf{S}$ $\in \mathbb{R}^{M \times d_{model}}$ containing information from $M$ channels, we have: 
\begin{equation}
    \mathbf{O} = \mathtt{SensorMixing}(\mathbf{S})
\end{equation}
where $\mathbf{O}$ $\in \mathbb{R}^d_{mix}$. We implement $\mathtt{SensorMixing}$ with a global average pooling along the channel dimension.
Finally, a $\mathtt{Linear}$ layer is added to get the predicted category of the IMU signal.
\begin{equation}
    \hat{y_i} = \mathtt{Softmax}(\mathtt{Linear}(\mathbf{O}))
\end{equation}
where $\hat{y_i}$ $\in \mathbb{R}^K$ is the vector representing the probabilities of $K$ activity categories for the IMU sample $\mathbf{X}$.

\subsection{Computational Efficiency Analysis}
\label{sec:efficiency}
\begin{figure}
\centering
\includegraphics[width=2.6in]{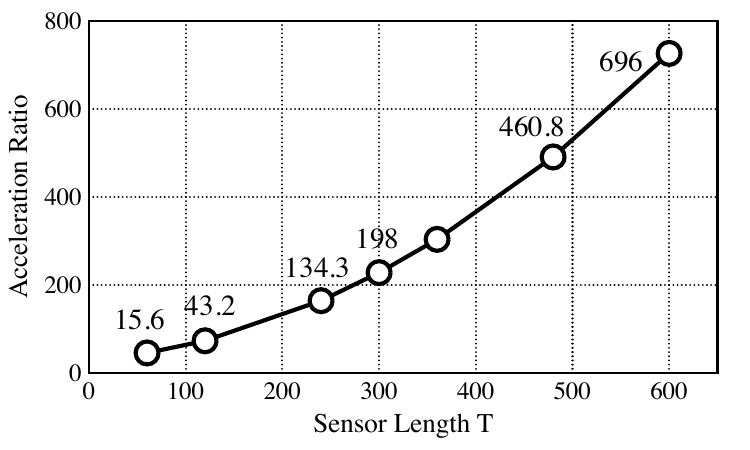}
\vspace{-6pt}
\caption{Acceleration ratio increases with the sensor data length.}
\label{fig:speedup_ratio}
\vspace{-6pt}
\end{figure}

In transformer-based models, the calculation of self-attention is a major component of the overall computation complexity~\cite{bert,transformer}. 
Given our novel design of sensor-wise attention, we reduce the computation complexity of self-attention significantly, which consists of the calculation of \textit{(i) $Q$ $K$ $V$ matrices} and \textit{(ii) the matrix multiplication} between $Q$ with $K$, and between the attention map with $V$. 
The main reduction of computation comes from two perspectives, i.e., sensor-wise attention and selective masking. 
By combining our sensor-wise attention and selective masking, we derive the total acceleration ratio in computation in the self-attention module: 

\begin{equation}
    Acceleration_{ratio} = \frac{2\times T^2 + 3\times T \times d_{model}}{M\times((C+S-1) + 3\times d_{model})}
\end{equation}
In our setting, $M=6$, $C=3$, $S=2$, $d_{model}=64$, and $T$ equals 120, thereby our $Acceleration_{ratio}$ is 43.2. We also visualize the $Acceleration_{ratio}$ in Fig.~\ref{fig:speedup_ratio}.

\section{Evaluation}
\label{sec:evaluation}

We aim to explore the following research questions. 

\begin{itemize}[leftmargin=*]
    \item \textbf{RQ1 (Reliability)}: What is the performance of \N\ compared to state-of-the-art methods? 
    \item \textbf{RQ2 (Efficiency)}: What is the efficiency of \N?
    \item \textbf{RQ3 (Interpretability)}: Whether important components of \N\ contribute to the final performance?
    \item \textbf{RQ4 (Explainability)}: Why sensor-wise attention is effective?
    
\end{itemize}

\subsection{Evaluation Setup}

\subsubsection{Evaluation metrics}

We utilize average accuracy and F1-score as evaluation metrics.

\subsubsection{Dataset and preprocessing}
Following previous studies~\cite{cross_person_aaai21,imubert,ubicomp20_har_da,unihar,ubicomp19_har_multitask_self,yao2017deepsense}, 
we utilize UCI~\cite{uci}, Shoaib~\cite{shoaib}, Motion~\cite{motionsense}, and HHAR~\cite{hhar} datasets. 
Each dataset is seen as one domain. 
All datasets contain four shared activities, 
i.e., \textit{walking, standing or sitting, walking upstairs, walking downstairs}. 
The sampling rate is 20, and the window size is 6. 
\textit{For source datasets}, we partition each dataset for training (80\%), validation (10\%) and testing (10\%). 
\textit{For target datasets}, the whole dataset is used for testing. 
All models are trained only with data from the source domain and are evaluated on target domains.

\subsubsection{Baselines}

We compare \N\ with 10 baseline models. 
These models are representative because they consist of pretraining-free models (MLP~\cite{baseline_mlp}, CNN~\cite{baseline_cnn}, ResNet~\cite{baseline_resnet}, LSTM~\cite{baseline_lstm}, and TSFCN~\cite{baseline_fcn}), pretraining models (LIMUBERT~\cite{imubert} and ContraTSC~\cite{tstcc_pami}), one SoA frequency learning method (FreHAR~\cite{frequency_mlp_nips23}), one SoA IMU domain generalization method (SDMix~\cite{cross_dataset_ubicomp22_jindong}), and one SoA IMU domain adaptation method (UniHAR~\cite{unihar}). 


\begin{itemize}[leftmargin=*]
    \item \textbf{MLP}~\cite{baseline_mlp} has recently received growing attention from time series~\cite{chen2023tsmixer} and computer vision~\cite{MLP_mixer} communities due to its simplicity and good performance in various tasks.   
    \item \textbf{CNN}~\cite{baseline_cnn} is widely applied for time series analysis and computer vision to extract features from data. 
    \item \textbf{ResNet}~\cite{baseline_resnet} (Residual Neural Network) is a popular deep learning model and it has achieved good performance in various tasks.
    \item \textbf{LSTM}~\cite{baseline_lstm} is a type of recurrent neural network architecture and is commonly used for sequence analysis. 
    \item \textbf{TSFCN}~\cite{baseline_fcn} (Time Series Fully Convolutional Network) has achieved very good performance in time series classification. The global average pooling layer is designed to identify the contribution region from raw IMU data.
    \item \textbf{LIMUBERT}~\cite{imubert} designs a self-supervised learning model to improve performance in activity recognition. The main component is the BERT-style pertaining task to learn useful IMU representation. 
    \item \textbf{ContraTSC}~\cite{tstcc_pami} employs contrastive learning for pre-training to learn robust representation for time series classification. It achieved good performance in multiple downstream time series classification tasks.
    \item \textbf{FreHAR}~\cite{frequency_mlp_nips23} is a state-of-the-art frequency learning model designed for time series data. It has achieved outstanding performance in multiple tasks. 
    \item \textbf{SDMix}~\cite{cross_dataset_ubicomp22_jindong} is a state-of-the-art cross-domain HAR model. The core design is a Semantic 
    Discriminative Mixup approach that overcomes the semantic inconsistency brought by domain gaps across a single domain or across datasets. 
    \item \textbf{UniHAR}~\cite{unihar} is a state-of-the-art cross-dataset HAR model. UniHAR is featured by a group of physics-informed IMU augmentation approaches that improve the generalization capability of HAR.

\end{itemize}

\subsubsection{Implementation details} 
Our framework was developed on a system equipped with 40GB of RAM and a P40 GPU with 24GB of memory. All models were implemented with Python 3.8 and PyTorch 1.12. 
The batch size is set to 512 and the number of training epochs to 150, with a learning rate of 0.001.

\begin{table*}[t]
\footnotesize
\centering
\caption{Overall cross-domain accuracy. The \textbf{bold} and \underline{underline} represent the best and second-best results.}
\vspace{-6pt}
\begin{tabular}{l|lcccccccccccc}
\hline
Source  & Target     & MLP & CNN &  ResNet & LSTM & TSFCN & LIMUBERT  & ContraTSC & FreHAR & SDMix & UniHAR  & \textbf{Our \N} & Improvement          \\ \hline
\hline
\multirow{3}{*}{UCI} &  Shoaib &  46.74  & 49.38 & 50.51 & 32.5 & 36.45   & 58.28 & 51.01 & 33.15 & 57.67 & \underline{72.13}  & \textbf{82.49} & 10.36\% \\ 
&  Motion &  53.08  & 43.68 & 52.49 & 64.75 & 46.19    & 68.94 & 52.29 & 69.84 & 66.18 & \underline{72.71} & \textbf{90.63}  & 17.92\%      \\
&  HHAR      & 45.94 & 56.46 & 51.21 & 68.15& 44.81  & 69.44 & 42.86 & 72.19 & 61.10 & \underline{74.66} & \textbf{81.67} & 7.01\%  \\ \hline
\multirow{3}{*}{Shoaib} &  UCI &  60.64  & 64.90 & 66.53 & 59.5 & 70.37   & 53.00 & 55.36 & 73.15  & 71.77 &  \underline{81.87} & \textbf{87.91}   & 6.04\%     \\ 
&  Motion &  58.59  & 72.05 &  59.18 & 73.78 & 60.03    & 69.50 & 53.55 & 80.60 &  65.07& \underline{85.62} &\textbf{90.9} & 5.28\%       \\
& HHAR      & 50.18 & 59.50 & 58.56 &  57.45 & \underline{60.57}  & 53.00 & 42.86 & 54.71 & 58.31 & 60.45 &  \textbf{65.7} & 5.13\%  \\ \hline
\multirow{3}{*}{Motion} &  UCI &  56.43  & 50.06 & 63.52 & 53.31 & 57.51   & 72.26 & 45.05 &  58.27 & 65.09 & \underline{72.50}  & \textbf{91.29}   & 18.79\%     \\ 
&  Shoaib &  57.21  & 56.92 & 60.37 & 56.82 & 55.74    & 65.72 & 43.21 & 69.96 & 59.93 & \underline{79.06} &\textbf{86.89}  & 7.83\%      \\
&  HHAR      & 52.37 & 43.12 & 53.49 & 43.7 & 49.62  & 64.63 & 42.86 & 53.03 & \underline{74.09} & 60.78  &\textbf{80.7} & 6.61\%  \\ \hline
\multirow{3}{*}{HHAR} & UCI &  54.51  & 48.85 & 48.02 & 40.5 & 48.02   & 67.04 & 55.54 & 68.58 & 54.49 & \underline{76.78}  & \textbf{81.03}    & 4.25\%    \\ 
&  Shoaib &  53.89  & 24.39 & 44.88 & 19.45 & 41.14    & 53.56 & 26.24 & 27.83 &\underline{57.49} & 56.62 & \textbf{66.25}  & 8.76\%      \\
&Motion      & 60.28 & 60.86 & 51.64 & 37.11 & 64.38  & 57.04 & \underline{65.87} & 65.65 & 61.14 & 64.18 &\textbf{71.59} & 5.72\%  \\ \hline
\rowcolor{lightgray}
Average  & - & 54.16 & 52.51 & 55.03 & 50.59 & 52.90  & 62.70 & 48.06 & 60.58 & 62.70 & \underline{71.45} &  \textbf{81.42} & 9.97\%  \\ \hline
\hline
\end{tabular}
\label{tab:over_eval}
\end{table*}

\subsection{Main Results (RQ1)}
In Table~\ref{tab:over_eval}, we present the overall cross-domain performance of \N\ compared with SoA methods. 
Note that for each model, we train them on four source datasets separately and evaluate them on corresponding target datasets. 
We can observe that the average accuracy of \N\ is 81.42\%.  
The second-best model is UniHAR, which yielded an average accuracy of 71.45\%. 
The third-best models are SDMix and LIMUBERT, which benefit from data augmentation and unsupervised pretraining. These results demonstrate the effectiveness of IMU data augmentation and pretraining in cross-domain HAR. 
For other baseline models, the performance is not good enough (average accuracy is less than 55\%) because they cannot capture the domain-invariant IMU representation by training on source domain data. 
Overall, our framework achieved the best performance across all datasets and outperformed the second-best model UniHAR by 9.97\% in average accuracy, demonstrating strong generalization capabilities. 
Due to the space limit, we present the average accuracy and omit the F1 score. 

\subsection{Efficiency Analysis for Deployment (RQ2)}
We compare the efficiency of \N\ with the three best-performing baseline models as shown in Table~\ref{table:efficiency}.  
FLOPs is an important metric since it strongly correlates with the energy consumption of deep learning models~\cite{onebit_llm}. 
We can observe that \N\ yields significantly smaller FLOPs ($52k$) compared to other methods ($> 300k$) with similar inference time. 
This property is extremely beneficial for large-scale industrial deployment since it requires less computation and energy costs from mobile devices.

\begin{table}[H]\footnotesize
\caption{Efficiency comparison.}
\vspace{-6pt}
\label{table:efficiency}
\begin{tabular}{lcccc}
\hline
Model     &  Parameter & FLOPS &    Inference Time (s)           \\ \hline
LIMUBERT &  67k & 1136k &    0.4089  \\
SDMix      & 205k& 530k  & 0.0140   \\
UniHAR     & 15k & 333k & 0.0362  \\  
\N\ (Ours)      & 25k & 52k &  0.0299  \\  \hline 
\end{tabular}
\end{table}

\subsection{Ablation Study (RQ3)}

\begin{figure}[t]
\centering
\includegraphics[width=3.3in]{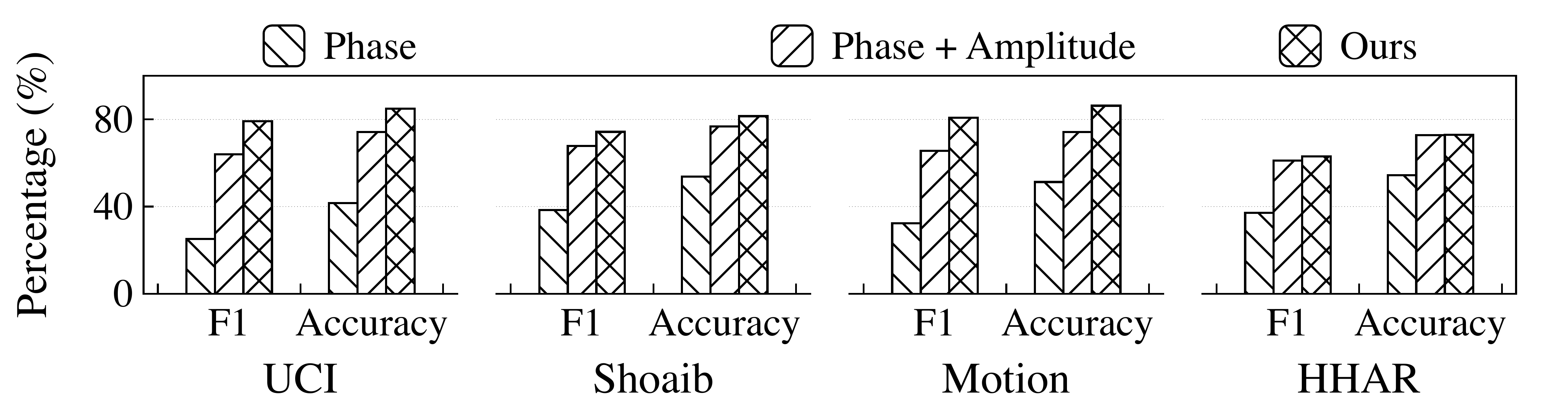}
\vspace{-6pt}
\caption{Impact of frequency amplitude.}
\label{fig:ablation_phase}
\vspace{-6pt}
\end{figure}

\vspace{3pt}
\noindent 
\textbf{Importance of frequency amplitude.} We design two variations of \N\ to evaluate the importance of amplitude and phase. Results are shown in Fig.~\ref{fig:ablation_phase}. 
\textit{(i) replace the amplitude with phase information}. Firstly, from Fig.~\ref{fig:ablation_phase}, we can see that the performance drops significantly without amplitude information. On average, the accuracy drops from 81.42\% to 50.23\% (drops by 31.19\%), and the F1 drops from 74.35\% to 33.22\% (drops by 41.13\%), which follows our intuition that amplitude reflects more important information for distinguishing different activities. It is challenging to learn domain-invariant representations from phase features. 
\textit{(ii) combine amplitude and phase information}. 
Secondly, combining amplitude and phase information also decreases the performance. On average, the accuracy drops from 81.42\% to 74.50\% (drops by 6.92\%), and the F1 drops from 74.35\% to 64.61\% (drops by 9.74\%). This observation follows our intuition that phase information might not benefit the generalization of HAR, as it is easily impacted by random noise.

\begin{figure}
\vspace{-6pt}
\centering
\includegraphics[width=3.3in]{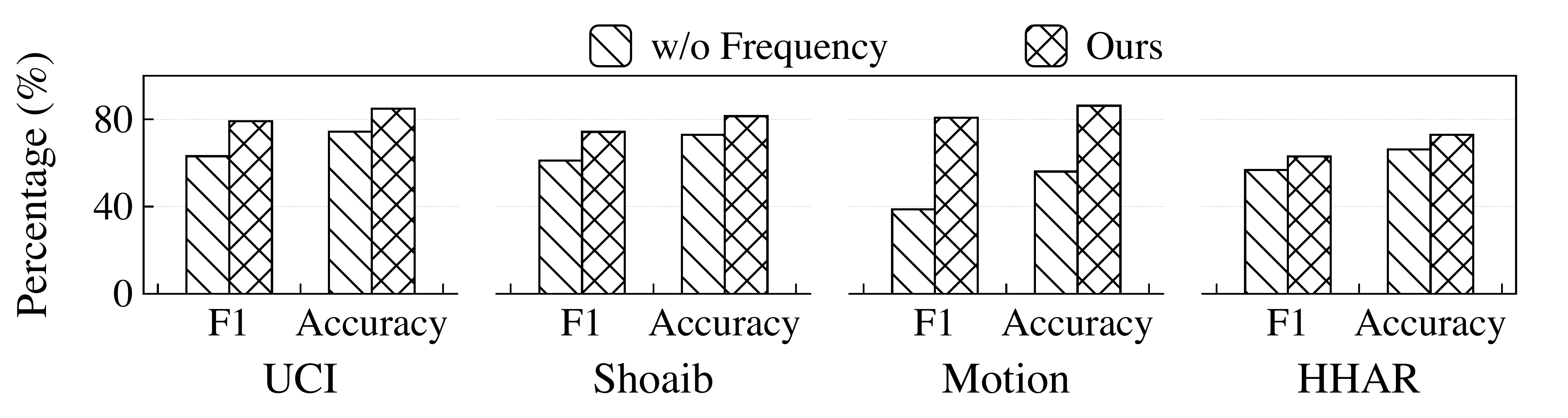}
\vspace{-6pt}
\caption{Impact of frequency features.}
\label{fig:ablation_time}
\vspace{-6pt}
\end{figure}

\vspace{3pt}
\noindent 
\textbf{Importance of frequency feature.} We replace frequency features with time features, i.e., temporal sensor sequences, and keep other components the same. 
Results are shown in Fig.~\ref{fig:ablation_time}. 
Note that for all datasets, \N\ achieved significantly better performance. 
On average, the accuracy drops from 81.42\% to 67.91\% (drops by 13.51\%), and the F1 drops from 74.35\% to 54.94\% (drops by 19.41\%), which demonstrates the great importance of frequency features.
These results demonstrate that \N\ learns more domain-invariant representations in the frequency space. 


\begin{figure}
\vspace{-6pt}
\centering
\includegraphics[width=3.3in]{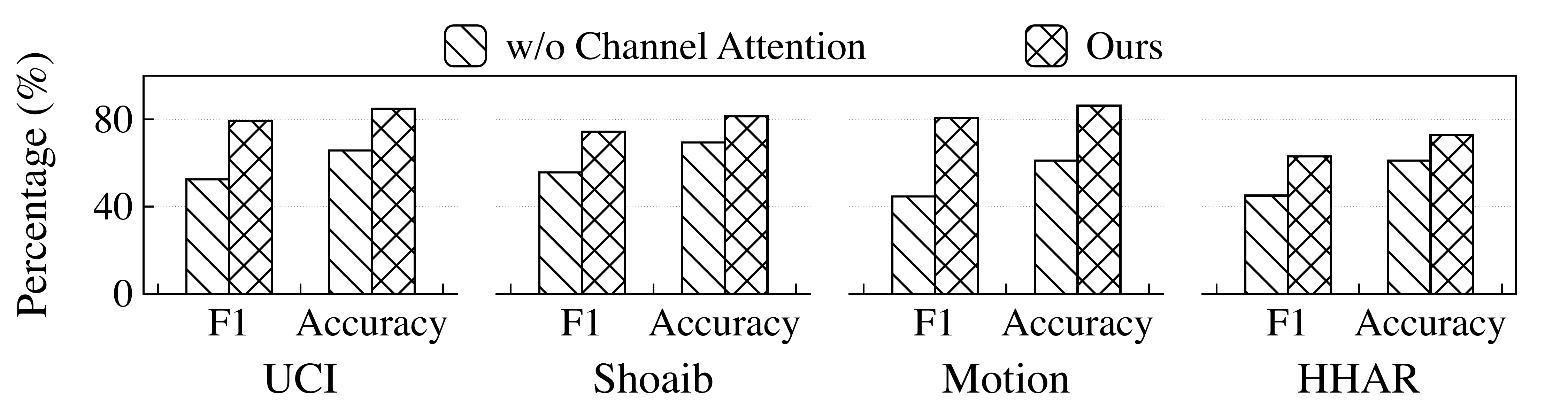}
\vspace{-6pt}
\caption{Impact of sensor-wise attention.}
\label{fig:ablation_channel_attention}
\vspace{-6pt}
\end{figure}

\vspace{3pt}
\noindent 
\textbf{Importance of sensor-wise attention.} We replace our sensor-wise attention with the commonly-used self-attention along the temporal dimension, i.e., the combination of all channels at each time step forms a token. 
Results are shown in Fig.~\ref{fig:ablation_channel_attention}. 
From the figure, we can see that the performance dropped on all datasets, especially on the Motion dataset. 
On average, the accuracy drops from 81.42\% to 64.32\% (drops by 17.10\%), 
which demonstrates that capturing the correlation among sensor channels benefits the generalization of HAR significantly.

\begin{figure}
\vspace{-6pt}
\centering
\includegraphics[width=3.3in]{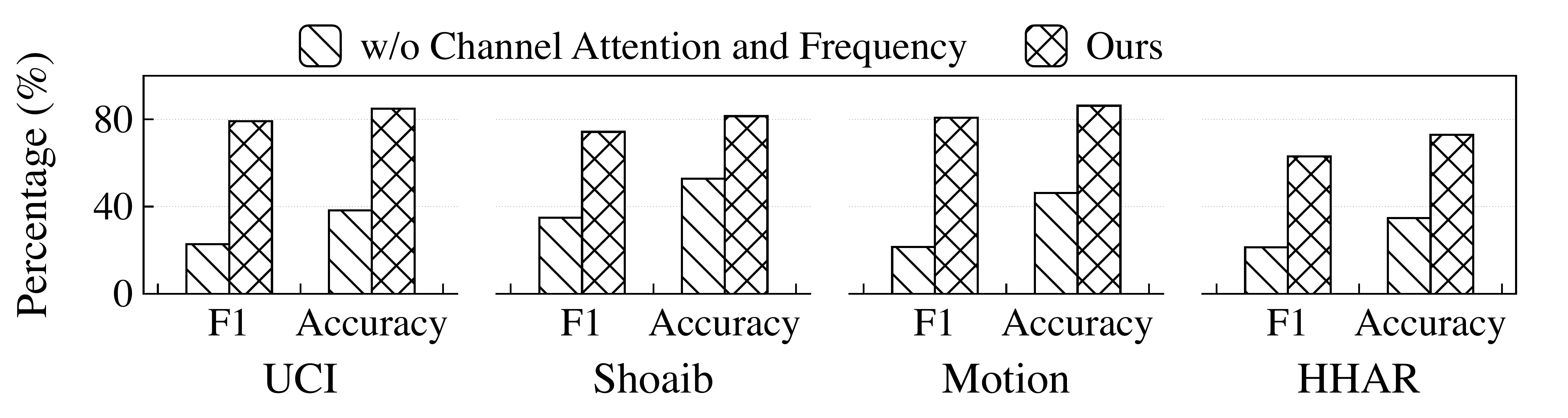}
\vspace{-6pt}
\caption{Impact of sensor-wise attention and frequency.}
\label{fig:ablation_channel_attention_freq_data}
\vspace{-6pt}
\end{figure}

\vspace{3pt}
\noindent 
\textbf{Importance of sensor-wise attention and frequency feature.} 
We replace both the frequency feature and the sensor-wise attention with a commonly-used self-attention along the temporal dimension on temporal features, i.e., raw IMU input. 
Results in Fig.~\ref{fig:ablation_channel_attention_freq_data} show that, on average, the accuracy drops from 81.42\% to 42.98\% (drops by 38.44\%), which demonstrates the great importance of sensor-wise attention and frequency features.

\begin{figure} 
\vspace{-6pt}
\begin{minipage}[c]{0.235\textwidth}
\includegraphics[width=\textwidth, keepaspectratio=true]{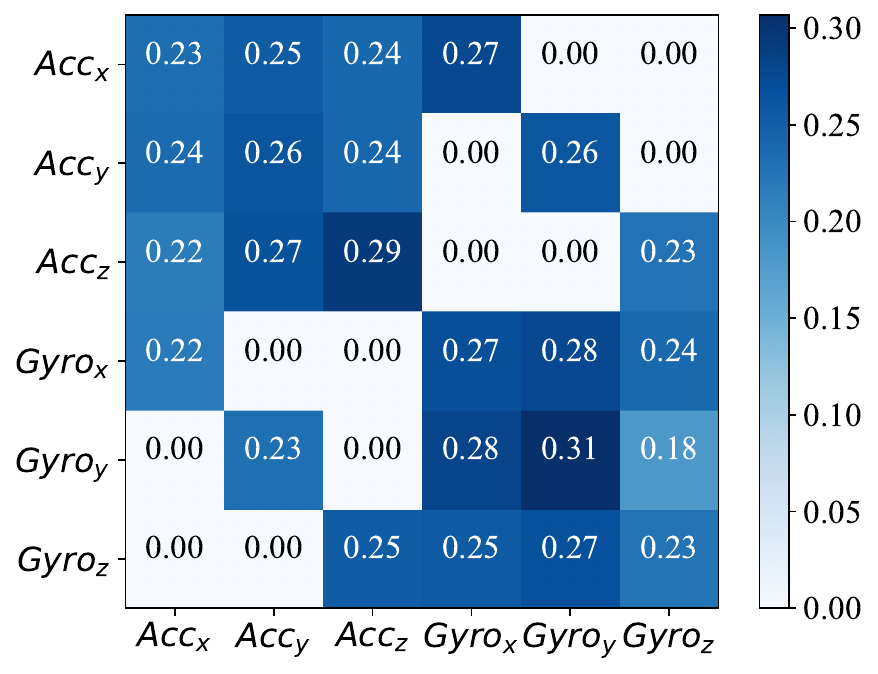}
\subcaption{Sensor-wise attention}
\label{fig:attention_map_uci_to_shoaib}
\end{minipage}
\begin{minipage}[c]{0.225\textwidth}
\includegraphics[width=\textwidth, keepaspectratio=true]{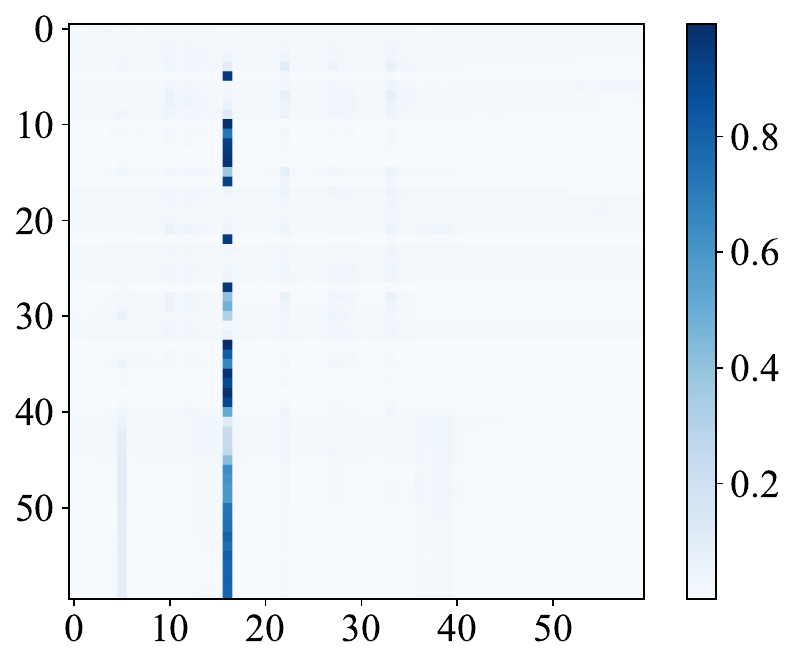}
\subcaption{Regular attention}
\label{fig:attention_map_uci_frequency_domain_frequency_self_atten_1}
\end{minipage}
\caption{Attention map comparison.}
\label{fig:attention_map}
\vspace*{-8pt}
\end{figure}

\subsection{Sensor-wise Attention Analysis (RQ4)} 
We further visualize the attention weights of our sensor-wise attention maps to explore why it has achieved good performance. 
As shown in Fig.~\ref{fig:attention_map_uci_to_shoaib}, each grid represents the correlation between two sensor channels within one IMU sample. For example, the grid on the top left represents the correlation between $Acc_x$ and itself. 
Those grids with a correlation of 0 are the ones masked by our selective masking mechanism. 
Overall, all channels have strong correlations with each other, which means that each channel can utilize the information in other channels. 
However, in Fig.~\ref{fig:attention_map_uci_frequency_domain_frequency_self_atten_1}, nearly all frequency components are strongly correlated only with the frequency near 20. This means that in frequency-wise attention, each frequency component cannot utilize the information in other components, thereby not generating robust representation.
This potentially explains the low performance compared to our sensor-wise attention as in Fig.~\ref{fig:ablation_channel_attention}.

\section{Deployment for Logistics Services} 
\label{sec:deployment}
We cooperate with JD Logistics, a leading logistics company in China, and deploy our framework to evaluate its positive impact on last-mile logistics services.

\subsection{Deployment}
Couriers use Personal Digital Assistants (PDAs), similar to smartphones, for parcel delivery. We deploy \N\ on couriers' PDAs to infer real-time activities. 

\begin{figure}[t]
\centering
\includegraphics[width=3.3in]{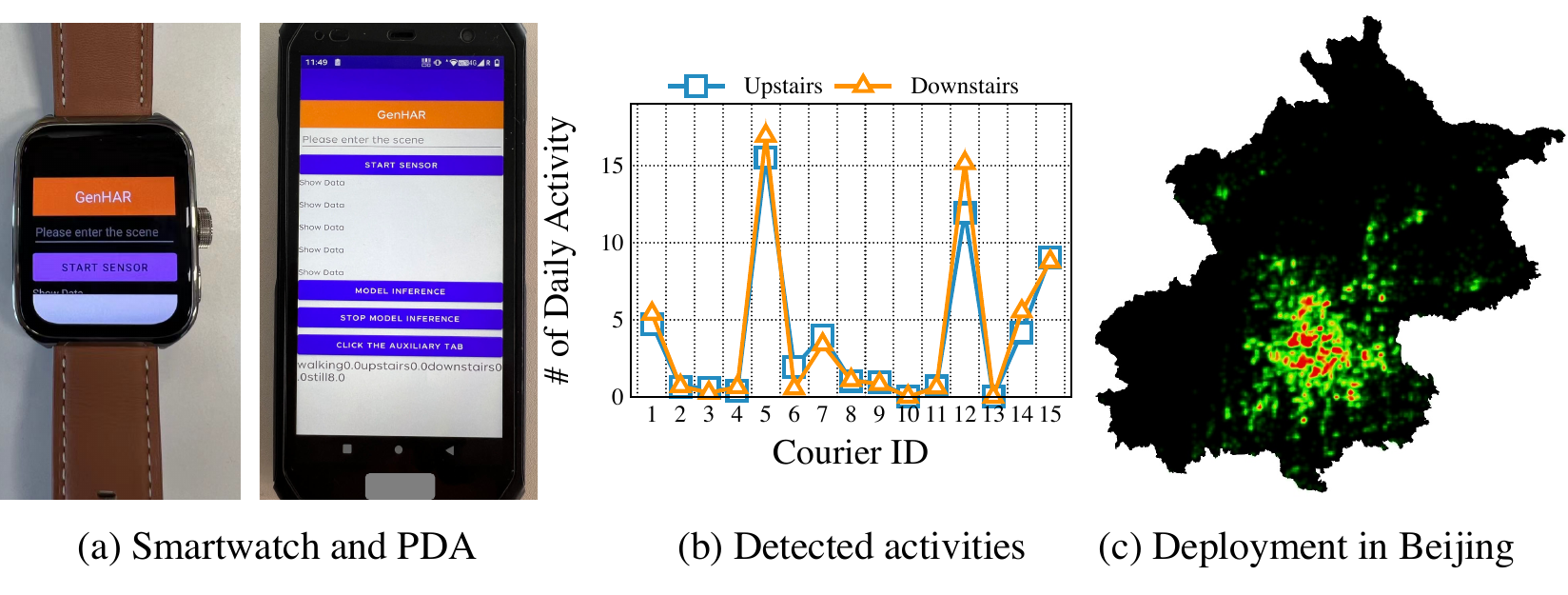}
\vspace{-6pt}
\caption{Illustration of real-world deployment results.}
\label{fig:deployment}
\vspace{-6pt}
\end{figure}

\vspace{2pt}
\noindent
\textbf{(i) Efficiency.} 
Before deploying \N\ on couriers' PDAs, we evaluate the efficiency by deploying \N\ on a regular PDA with the Android 11 system as shown in Fig.~\ref{fig:deployment}(a). 
The \textit{pt} file of \N\ is 183KB. 
We further evaluate the inference speed by running \N\ on the PDA 100 times. 
The average time taken was 0.00647 seconds for \N\ to make an activity recognition, with a minimum and maximum of 0.004 seconds and 0.012 seconds, respectively. 
We also evaluate the efficiency of \N\ on smartwatches (OppoWatch), the average inference time is 0.02505 seconds, with a minimum and maximum of 0.007 and 0.067 seconds, respectively. 


\vspace{2pt}
\noindent
\textbf{(ii) Effectiveness.} To avoid disturbing the normal working of couriers, we cannot ask them to manually label the activity ground truth. Therefore, we record their activity ground truth during parcel delivery by following 6 delivery couriers in Beijing City. 
The average accuracy for walking, upstairs, downstairs, and still are 82.4\%, 54.5\%, 88.1\%, and 99.8\%, respectively. 
The upstairs activity is hard to recognize, the downstairs, however, is accurate and thereby can support logistics services such as courier workload measurement.

\vspace{2pt}
\noindent
\textbf{(iii) Deployment scale.} 
We deployed \N\ on the mobile devices of \courier\ couriers in 4 Chinese cities. 
In one month, \N\ has detected \all\ activities, including \still\ still, \walking\ walking, \up\ upstairs, and \down\ downstairs.

\subsection{Applications}

\vspace{2pt}
\noindent
\textbf{(i) Delivery behavior understanding.} 
We deploy \N\ on mobile devices (PDAs) carried by couriers for understanding their delivery-related activities. 
We take the delivery region with 15 couriers in Beijing City as an example and show the detected results in Fig.\ref{fig:deployment}(b). 
By combining couriers' behaviors, such as upstairs and walking, with delivery log data, we can systematically identify high-effort delivery scenarios. 
For example, buildings without elevators that require significantly more stair climbing (e.g., courier 5 in Fig.\ref{fig:deployment}(b)), or residential areas that prohibit delivery vehicles, bring more walking to reach delivery points. 
Such delivery behavior understanding can be used to support more equitable compensation policies (e.g., providing additional incentives for high-effort deliveries). 
Also, it can be used for more fair and workload-aware dispatching algorithms~\cite{santong}, improving both efficiency and fairness in last-mile delivery.

\vspace{2pt}
\noindent
\textbf{(ii) Address location mining.} 
Accurate locations of shipping addresses are critical for logistics applications~\cite{singhal2024geoindia,gis_baidu,cominer}. 
Since delivery couriers visit shipping addresses every day, we leverage the detected courier activity data and corresponding GPS trajectories to infer shipping address coordinates. 
Specifically, we utilize the detected activity results from \N\ to determine whether the reported coordinates from couriers are reliable. 
For example, if the courier is driving continuously within a 5-minute window around the reporting time, the reported coordinates could be discarded to improve the inference of address locations. 
Finally, using activity recognition results, our system has successfully inferred the locations of over 6.8 million shipping addresses. 
We show an example of inferred addresses in Beijing City in Fig.~\ref{fig:deployment}(c). 
With 1,000 ground truth locations, the average accuracy of the inferred locations is 93.3\% for 100-meter accuracy, a widely adopted metric for shipping address location accuracies~\cite{cominer}. 
Inferred address locations with such high accuracy benefit logistics platforms for navigation, parcel allocation, etc.


\section{Discussions}
\label{sec:discussion}
In this section, we discuss the lessons learned, our limitations, and potential future directions. 
\subsection{Lessons Learned}
\begin{itemize}[leftmargin=*]
    \item \textbf{Frequency space for domain generalization.} Our experiments have shown that representing sensor data in the frequency domain reduces the gap between different domains. This finding suggests that training models on one domain and applying them to another becomes more effective when considering the frequency domain. This discovery has significant implications for future research and applications in time-series classification problems based on sensor data. 
    \item \textbf{Physical-driven model design improves performance.} We design the selected attention mechanism motivated by the physical generation principles of IMU signals. Empirical results suggest this design improves the generalization capability significantly. This finding can be extended to areas where the data generation process can be formulated by physical models, such as WiFi. 
\end{itemize}

\subsection{Limitations and Future Work}
\label{app:limitations}
While \N\ achieved good performance in generalized human activity recognition, there are still limitations we can address to further improve the performance. 
Firstly, even though our design is highly efficient and requires much fewer FLOPs compared to SoA methods (see Table~\ref{table:efficiency}), we identify potential directions that can further reduce the parameters and FLOPs. For example, by cutting off high-frequency components, we can reduce the input length and parameters. 

In future studies, we are interested in extending our framework to other real-time mobile sensing tasks. For example, real-time anomaly detection and hand gesture recognition, utilizing the high-efficiency advantage of \N.  
Also, we will further improve the efficiency by approaches such as parameter sharing and model quantification~\cite{onebit_llm}.

\vspace{-3pt}
\section{Related Work}
\label{sec:related_work}

\subsection{Human Activity Recognition}
With the widely deployed mobile devices such as smartwatches, the ubiquitous sensors on these devices provide a unique opportunity for human activity recognition~\cite{ubicomp19_har_toolkit,ubicomp21_har_crowdact,ubicomp20_har_da,ubicomp22_har_katn,ubicomp20_har_personalization,ubicomp22_har_distengle,ubicomp23_har_boost,ubicomp20_har_multitask,ubicomp21_har_augmentation,gao2023mmtsa}.
Chen et al.~\cite{survey_har_yunhao} and Cai et al.~\cite{cai_survey} provide a systematic review of sensor-based human activity recognition. 
To increase the safety and robustness of activity recognition models, 
Jeyakumar et al.~\cite{ubicomp23_har_explain} design an explainable Complex Human Activity Recognition model for critical applications such as healthcare.
Recently, with the increasing availability of unlabeled data, weak-supervised, self-supervised, and unsupervised methods are receiving more attention~\cite{ubicomp19_har_multitask_self,ubicomp20_har_weaksupervised,ubicomp22_har_colloss,ubicomp21_har_selfhar,ubicomp21_har_contrastive,ubicomp22_har_katn,ubicomp22_har_assessing,imubert}. 
Due to the heterogeneity of sensor readings, models trained on one domain might not achieve comparable performance on other domains~\cite{ubicomp19_har_cross_dataset,presotto2023combining,imubert,llm4har,deplpoyhar}. 
Therefore, some recent studies focus on the generalization capability of HAR models. 
Xu et al.~\cite{unihar} design an augmentation approach and utilize unlabeled target domain data to enhance cross-domain HAR.
Taking one step further, some methods have been recently designed to train only with data from source domains~\cite{cross_dataset_ubicomp22_jindong,TF-C_nips22}. 
Different from most studies, we focus on HAR without any labeled or unlabeled training data from the target domain.

\subsection{Frequency Learning}
Recently, the frequency space has received growing attention from the time series research community~\cite{frequency_ts_survey,frequency_channel_har_tkde23,DG_knowledge_distill_jindong,units_sensys21}. 
Most studies adapt frequency as a complementary source of feature, and design models to combine frequency and time features~\cite{he2023domain,yang2022unsupervised,liu2023temporal,TF-C_nips22,wu2021autoformer}.
At the same time, frequency alone to provide time series information is less explored~\cite{frequency_mlp_nips23,DG_knowledge_distill_jindong}.
Note that, while there is increasing attention in the frequency space, it is still an open question whether frequency could provide enough information, especially for IMU data analysis.
Moreover, we aim to learn domain-invariant representation purely in the frequency space, which is significantly different from most existing studies that utilize frequency as complementary features or as a noise filter.

\vspace{-3pt}
\section{Conclusion}
\label{sec:conclusion}

In this study, we focus on cross-domain human activity recognition. 
To tackle the distribution shift challenges, we design \N, a novel framework for learning domain-invariant representation of IMU data. 
\N\ explores the frequency space and designs an efficient sensor-wise self-attention mechanism. 
Systematic evaluation demonstrates that \N\ outperforms SoA methods significantly with much fewer computational costs. 
\N\ has been deployed in the real world to support last-mile delivery services.

\begin{acks}
We thank all the reviewers for their insightful feedback to improve this paper. This work is partially supported by the Financial Support for Outstanding Scientific and Technological Innovation Talents Training Fund in Shenzhen, and 
This research is partially supported by the National Research Foundation
(NRF), Prime Minister’s Office, Singapore, under its Campus for
Research Excellence and Technological Enterprise (CREATE) programme. The Mens, Manus, and Machina (M3S) is an interdisciplinary research group (IRG) of the Singapore-MIT Alliance for
Research and Technology (SMART) centre.
\end{acks}


\bibliographystyle{ACM-Reference-Format}
\bibliography{survey}


\end{document}